\documentclass[conference]{IEEEtran}
\IEEEoverridecommandlockouts
\usepackage{comment}
\usepackage{amsmath,amssymb,amsfonts}
\usepackage{algorithmic}
\usepackage{graphicx}
\usepackage{textcomp}
\usepackage{xcolor}
\usepackage{graphicx}
\usepackage{amsmath}
\usepackage{cite}
\usepackage{wrapfig}
\usepackage{soul}

\usepackage[colorlinks,urlcolor=blue,linkcolor=blue,citecolor=blue]{hyperref}

\usepackage{mdframed}
\usepackage{tcolorbox}
\def\BibTeX{{\rm B\kern-.05em{\sc i\kern-.025em b}\kern-.08em
    T\kern-.1667em\lower.7ex\hbox{E}\kern-.125emX}}
\begin{document}

\title{Neurosymbolic AI - Why, What, and How}
\author{Anonymous Authors}
\author{\and
\IEEEauthorblockN{\textbf{Amit Sheth}}
\IEEEauthorblockA{Artificial Intelligence Institute\\
University of South Carolina\\
Columbia, SC, USA \\
\texttt{amit@sc.edu}}
\and
\IEEEauthorblockN{\textbf{Kaushik Roy}}
\IEEEauthorblockA{Artificial Intelligence Institute\\
University of South Carolina\\
Columbia, SC, USA \\
\texttt{kaushikr@email.sc.edu}}
\and
\IEEEauthorblockN{\textbf{Manas Gaur}}
\IEEEauthorblockA{University of Maryland\\
Baltimore County\\
MD, USA \\
\texttt{manas@umbc.edu}}
}
\maketitle
\begin{abstract}
    Humans interact with the environment using a combination of perception - transforming sensory inputs from their environment into symbols, and cognition - mapping symbols to knowledge about the environment for supporting abstraction, reasoning by analogy, and long-term planning. Human perception-inspired machine perception, in the context of AI, refers to large-scale pattern recognition from raw data using neural networks trained using self-supervised learning objectives such as next-word prediction or object recognition. On the other hand, machine cognition encompasses more complex computations, such as using knowledge of the environment to guide reasoning, analogy, and long-term planning. Humans can also control and explain their cognitive functions. This seems to require the retention of  symbolic mappings from perception outputs to knowledge about their environment. For example, humans can follow and explain the guidelines and safety constraints driving their decision-making in safety-critical applications such as healthcare, criminal justice, and autonomous driving. While data-driven neural network-based AI algorithms effectively model machine perception, symbolic knowledge-based AI is better suited for modeling machine cognition. This is because symbolic knowledge structures support explicit representations of mappings from perception outputs to the knowledge, enabling traceability and auditing of the AI system’s decisions. Such audit trails are useful for enforcing application aspects of safety, such as regulatory compliance and explainability, through tracking the AI system's inputs, outputs, and intermediate steps. This first article in the Neurosymbolic AI department introduces and provides an overview of the rapidly emerging paradigm of Neurosymbolic AI, combining neural networks and knowledge-guided symbolic approaches to create more capable and flexible AI systems. These systems have immense potential to advance both algorithm-level (e.g., abstraction, analogy, reasoning) and application-level (e.g., explainable and safety-constrained decision-making) capabilities of AI systems.   
\end{abstract}
\section{Why Neurosymbolic AI?}
Neurosymbolic AI refers to AI systems that seek to integrate neural network-based methods with symbolic knowledge-based approaches. We present two perspectives to understand the need for this combination better: (1) algorithmic-level considerations, e.g., ability to support abstraction, analogy, and long-term planning. (2) application-level considerations in AI systems, e.g., enforcing explainability, interpretability, and safety.

\textbf{Algorithm-Level Considerations}

Researchers have identified distinct systems in the human brain that are specialized for processing information related to perception and cognition. These systems work together to support human intelligence and enable individuals to understand and interact with the world around them. Daniel Kahneman popularized a distinction between the goals and functions of  System 1 and System 2 \cite{kahneman2011thinking}. System 1 is crucial for enabling individuals to make sense of the vast amount of raw data they encounter in their environment and convert it into meaningful symbols (e.g., words, digits, and colors) that can be used for further cognitive processing. System 2 performs more conscious and deliberative higher-level functions (e.g., reasoning and planning). It uses background knowledge to position the perception module's output accurately, enabling complex tasks such as analogy, reasoning, and long-term planning. Despite having different functions, Systems 1 and 2 are interconnected and collaborate to produce the human experience. Together, these systems enable people to see, comprehend, and act, following their knowledge of the environment. 
In the past decade, neural network algorithms trained on enormous volumes of data have demonstrated exceptional machine perception, e.g., high performance on self-supervision tasks such as predicting the next word and recognizing digits. Remarkably, training on such simple self-supervision tasks has led to impressive solutions to challenging problems, including protein folding, efficient matrix multiplication, and solving complex puzzles \cite{jumper2021highly,fawzi2022discovering}.  However, knowledge enables humans to engage in cognitive processes beyond what is explicitly stated in available data. For example, humans make analogical connections between concepts in similar abstract contexts through mappings to knowledge structures that spell out such mappings \cite{gentner1983structure}. 
Perhaps current generative AI systems such as GPT-4 can acquire the  knowledge structures to support cognitive functionality from data alone \cite{bubeck2023sparks}. The hypothesis is that next-word prediction from many texts on the Internet can lead to an emergent  'cognitive model' of the world that the neural network can use to support cognition. However, significant concern regarding their black-box nature and the resulting inscrutability hinders the reliable evaluation of their cognitive capabilities. On the other hand, though unsuited for high-volume data processing, a symbolic model is highly suited for supporting human-like cognition using knowledge structures (e.g., knowledge graphs). Thus, rather than depend on one system or the other, it makes more sense to integrate the two types of systems: neural network-based Systems 1, adept at big-data-driven processing, and symbolic knowledge-based Systems 2, adept at dealing with knowledge-dependent cognition.

\textbf{Application-Level Considerations}

The combination of Systems 1 and 2 in Neurosymbolic AI can enable important application-level features, such as explainability, interpretability, safety, and trust in AI. Recent research on explainable AI (XAI) methods that explain neural network decisions primarily involves post-hoc techniques like saliency maps, feature attribution, and prototype-based explanations. Such explanations are useful for developers but not easily understood by end-users. Additionally, neural networks can fail due to uncontrollable training-time factors like data artifacts, adversarial attacks, distribution shifts, and system failures. To ensure rigorous safety standards, it is necessary to incorporate appropriate background knowledge to set guardrails during training rather than as a post-hoc measure. Symbolic knowledge structures can provide an effective mechanism for imposing domain constraints for safety and explicit reasoning traces for explainability. These structures can create transparent and interpretable systems for end-users, leading to more trustworthy and dependable AI systems, especially in safety-critical applications \cite{sheth2022process}.
\begin{tcolorbox}[title=Why Neurosymbolic AI?]
\textit{Embodying intelligent behavior in an AI system must involve both perception - processing raw data, and cognition - using background knowledge to support abstraction, analogy, reasoning, and planning. Symbolic structures represent this background knowledge explicitly. While neural networks are a powerful tool for processing and extracting patterns from data, they lack explicit representations of background knowledge, hindering the reliable evaluation of their cognition capabilities. Furthermore, applying appropriate safety standards while providing explainable outcomes guided by concepts from background knowledge is crucial for establishing trustworthy models of cognition for decision support.}
\end{tcolorbox}

\section{What is Neurosymbolic AI and How do we Achieve it?}
\begin{figure*}[!htb]
    \centering
    \tcbox{\includegraphics[width=0.93\linewidth,trim = 0cm 0cm 0cm 0cm, clip]{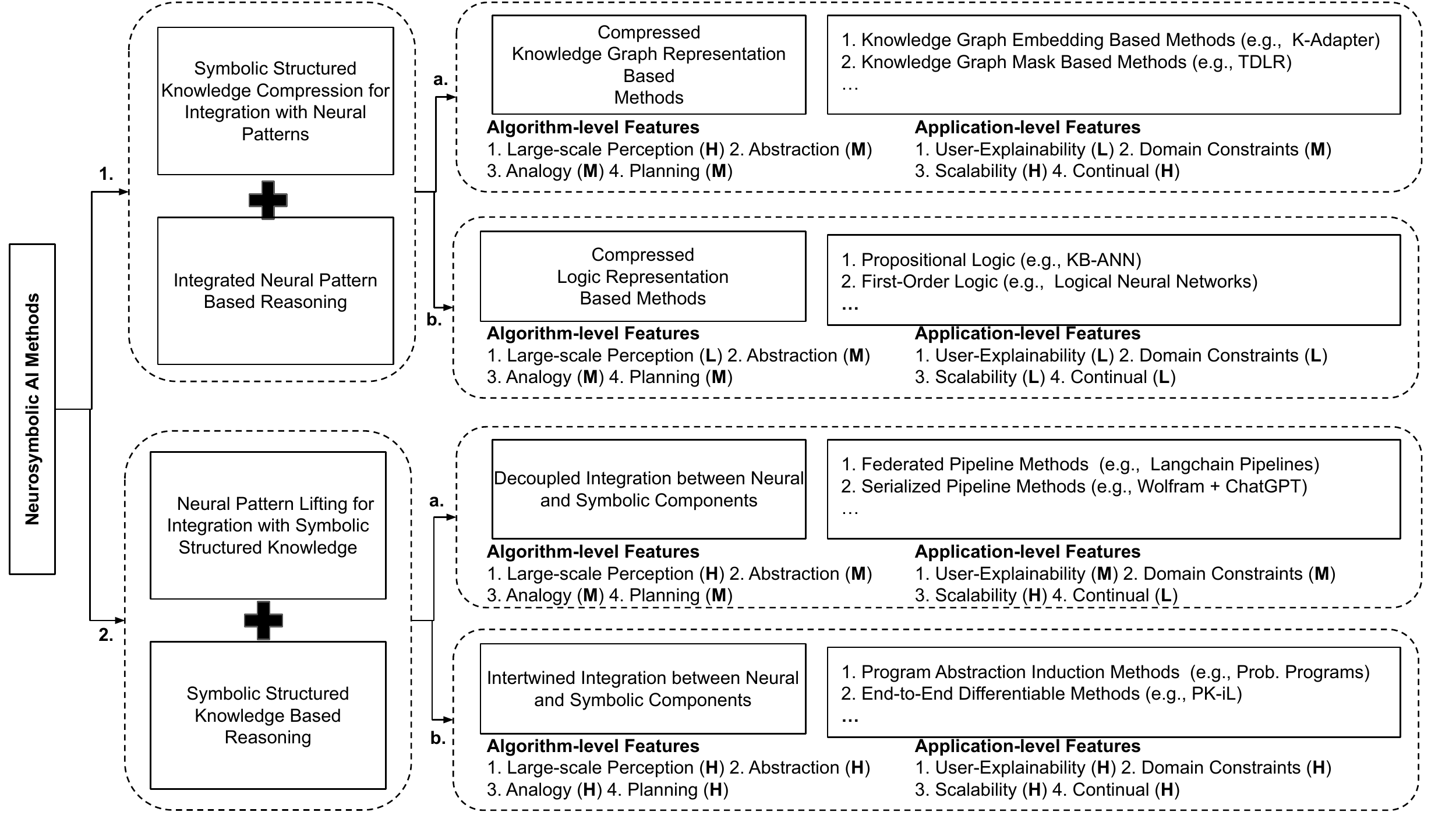}}
    \caption{\footnotesize The two primary types of neurosymbolic techniques—lowering and lifting—can be further divided into four sub-categories. Across the low (\textbf{L}), medium (\textbf{M}), and high (\textbf{H}) scales, these methods can be used to provide a variety of functions at both algorithmic and application levels.}
    \label{fig:landscape}
\end{figure*}

Neurosymbolic AI is a term used to describe techniques that aim to merge the knowledge-based symbolic approach with neural network methods to improve the overall performance of AI systems. These systems have the ability to blend the powerful approximation abilities of neural networks with the symbolic reasoning capabilities that enable them to reason about abstract concepts, extrapolate from limited data, and generate explainable results\cite{garcez2023neurosymbolic}. Together, these components support both algorithm-level and application-level concerns introduced in the previous sections. Neurosymbolic AI methods can be classified under two main categories: (1) methods that compress structured symbolic knowledge to integrate with neural patterns and reason using the integrated neural patterns and (2) methods that extract information from neural patterns to allow for mapping to structured symbolic knowledge (lifting) and perform symbolic reasoning. Furthermore, we sub-categorize (1) into methods that utilize (a) compressed knowledge graph representations for integration with neural patterns and (b) compressed formal logic-based representations for integration with neural patterns. We also sub-categorize (2) into methods that employ (a) decoupled integration between the neural and symbolic components and (b) intertwined integration between the neural and symbolic components. These methods enable both algorithm-level and application-level functions in varying degrees of effectiveness spanning low (\textbf{L}), medium (\textbf{M}), and high (\textbf{H}) scales. Figure \ref{fig:landscape} details our categorization of neurosymbolic AI methods.

\textbf{Algorithm-Level Analysis of Methods in Category 1.}

\begin{figure}[!htb]
    \centering
    \tcbox{\includegraphics[width=0.8\linewidth,trim = 0cm 3.5cm 0cm 0cm,clip]{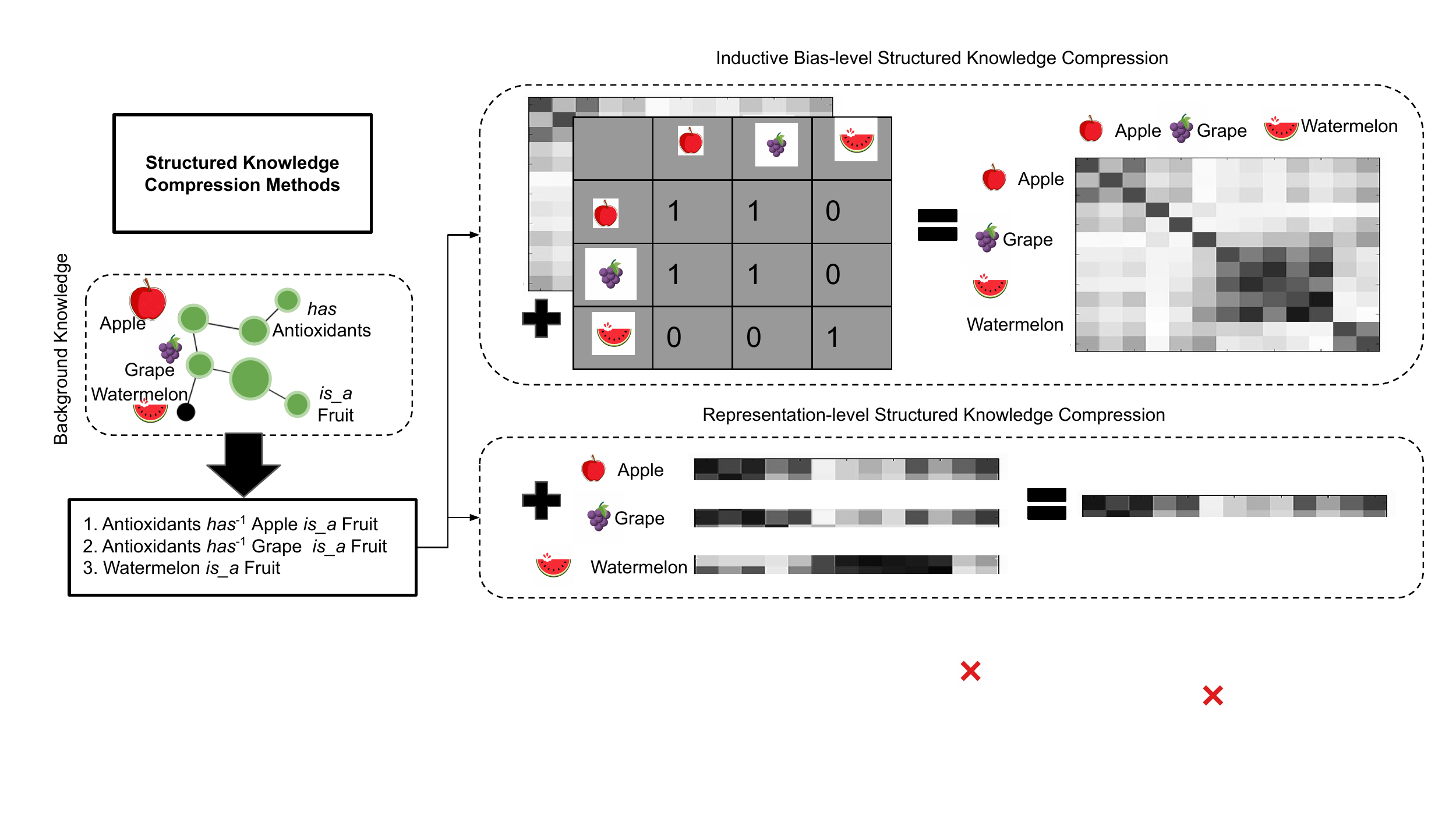}}
    \caption{\footnotesize The figure illustrates two methods for compressing knowledge graphs to integrate them with neural processing pipelines. One approach involves embedding knowledge graph paths into vector spaces, enabling integration with the neural network's hidden representations. The other method involves encoding knowledge graphs as masks to modify the neural network's inductive biases. An example of an inductive bias is the correlation information stored in the self-attention matrices of a transformer neural network \cite{rawte22tdlr,wang2020k}.}
    \label{fig:know_compress}
\end{figure}

For category 1(a), previous work has  used two methods to compress knowledge graphs. One approach is to use knowledge graph embedding methods, which compress knowledge graphs by embedding them in high-dimensional real-valued vector spaces using techniques such as graph neural networks. This enables integration with the hidden representations of the neural network. The other approach is to use knowledge graph masking methods, which encode the knowledge graphs in a way suitable for integration with the inductive biases of the neural network. Figure \ref{fig:know_compress} illustrates the two approaches. The ability of neural networks to process large volumes of raw data also translates to neural networks used for knowledge graph compression when processing millions and billions of nodes and edges, i.e., large-scale perception ((\textbf{H}) in Figure \ref{fig:landscape}). Utilizing the compressed representations in neural reasoning pipelines improves the system’s cognition aspects, i.e., abstraction, analogy, and planning capabilities. However, the improvements are modest ((\textbf{M}) in Figure \ref{fig:landscape}) due to the lossy compression of the full semantics in the knowledge graph (e.g., relationships aren’t modeled effectively in compressed representations). Category 1(b) methods use matrix and higher-order tensor factorization methods to obtain compressed representations of objects and formal logic statements that describe the relationships between them (such as propositional logic, first-order logic, and second-order situation calculus),  Improvements in cognition aspects follow a similar trend as in 1(a). However, compression techniques for formal logic are computationally inefficient and do not facilitate large-scale perception. ((\textbf{L}) in Figure \ref{fig:landscape}). 

\textbf{Application-Level Analysis of Methods in Category 1.}

For category 1(a), when compressing the knowledge graph for integration into neural processing pipelines, its full semantics are no longer explicitly retained. Post-hoc explanation techniques, such as saliency maps, feature attribution, and prototype-based explanations, can only explain the outputs of the neural network. These explanations are primarily meant to assist system developers in diagnosing and troubleshooting algorithmic changes in the neural network's decision-making process. Unfortunately, they are not framed in domain or application terms and hence have limited value to end-users ((\textbf{L}) for low explainability in Figure \ref{fig:landscape}). Knowledge graph compression methods can still be utilized to apply domain constraints, such as specifying modifications to pattern correlations in the neural network, as depicted in Figure \ref{fig:know_compress}. Nonetheless, this process has limited constraint specification capabilities, because large neural networks have multiple processing layers and moving parts ((\textbf{M}) in Figure \ref{fig:landscape}). It is challenging to determine whether modifications made to the network are retained throughout the various processing layers. Neural processing pipelines do offer a high degree of automation, making it easier for a system to scale across various use cases (such as plugging in use case-specific knowledge graphs) and to support continual adaptation throughout the system's life cycle (such as making continual modifications to the knowledge graphs). This capability is indicated by the letter (\textbf{H}) in Figure \ref{fig:landscape}. For category 1(b), when compressed formal logic representations are integrated with neural processing pipelines, system scores tend to be low across all application-level aspects of user-explainability, domain constraints, scalability, and continual adaptation, as denoted by the letter (\textbf{L}) in Figure \ref{fig:landscape}. This is primarily due to the effect of a significant user-technology barrier. End-users must familiarize themselves with the rigor and details of formal logic semantics to communicate with the system (e.g., to provide domain constraint specifications).

\textbf{Algorithm-Level Analysis of Methods in Category 2.}

\begin{figure*}[!htb]
    \centering
    \tcbox{\includegraphics[width=0.93\linewidth,trim = 0cm 0cm 5.5cm 0cm, clip]{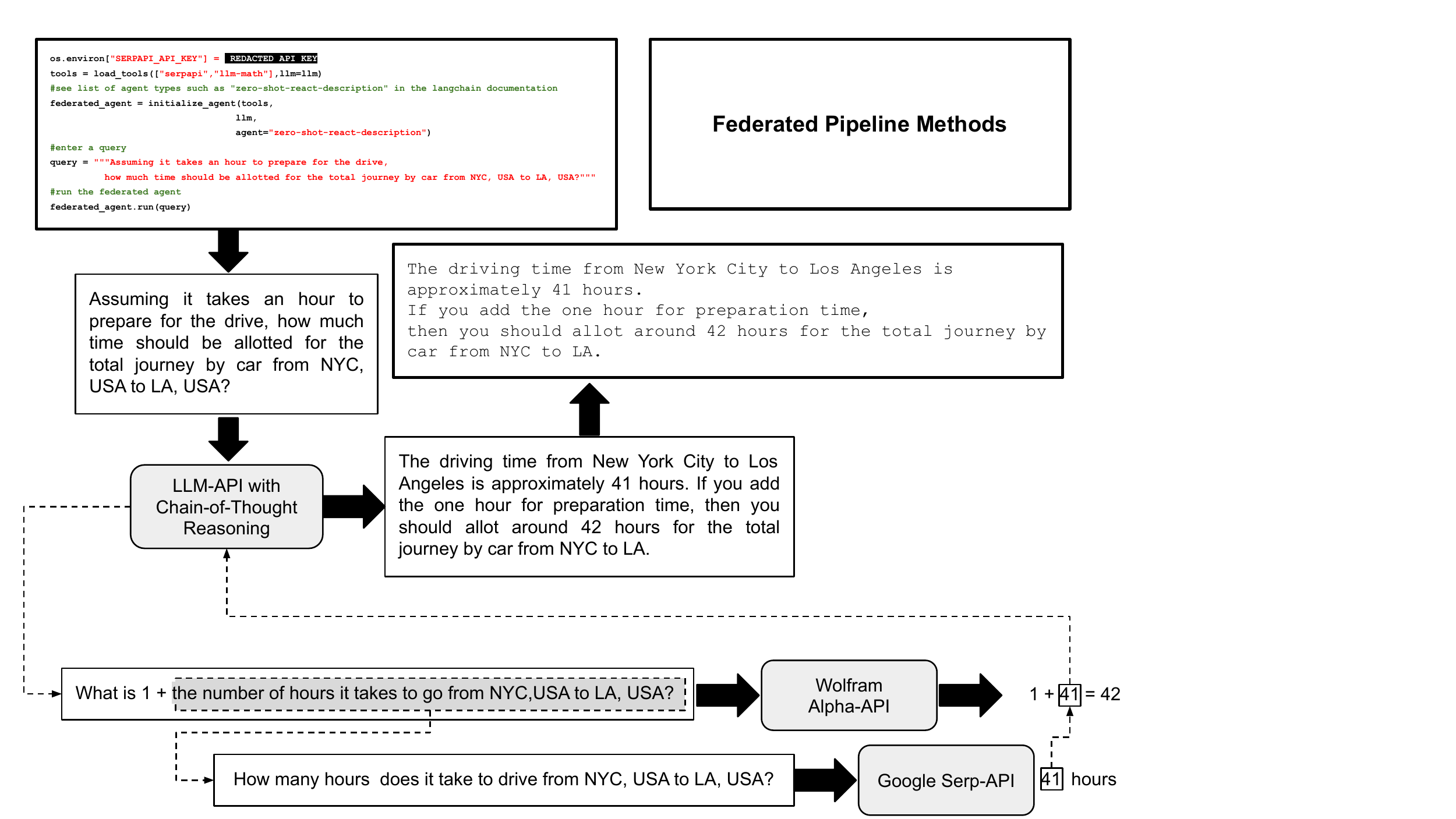}}
    \caption{\footnotesize Illustrates a federated pipeline method using the Langchain library. The method employs a language model trained on chain-of-thought reasoning to segment the input query into tasks. The language model then utilizes task-specific symbolic solvers to derive solutions. Specifically, the language model recognizes that search and scientific computing (mathematics) symbolic solvers are necessary for the given query. The resulting solutions are subsequently combined and transformed into natural language for presentation to the user.}
    \label{fig:two_a}
\end{figure*}

For category 2(a), the proliferation of large language models and their corresponding plugins has spurred the development of federated pipeline methods. These methods utilize neural networks to identify symbolic functions based on task descriptions that are specified using appropriate modalities such as natural language and images. Once the symbolic function is identified, the method transfers the task to the appropriate symbolic reasoner, such as a math or fact-based search tool. Figure \ref{fig:two_a} illustrates a federated pipeline method that utilizes the Langchain library. These methods are proficient in supporting large-scale perception through the large language model ((\textbf{H}) in Figure \ref{fig:landscape}). However, their ability to facilitate algorithm-level functions related to cognition, such as abstraction, analogy, reasoning, and planning, is restricted by the language model's comprehension of the input query ((\textbf{M}) in Figure \ref{fig:landscape}). Category 2(b) methods use pipelines similar to those in category 2(a) federated pipelines. However, they possess the added ability to fully govern the learning of all pipeline components through end-to-end differential compositions of functions that correspond to each component. This level of control enables us to attain the necessary levels of cognition on aspects of abstraction, analogy, and planning that is appropriate for the given application ((\textbf{H}) in Figure \ref{fig:landscape}) while still preserving the large-scale perception capabilities. Figure 4 shows an example of this method for mental health diagnostic assistance.

\begin{figure*}
    \centering
    \tcbox{\includegraphics[width=0.93\linewidth]{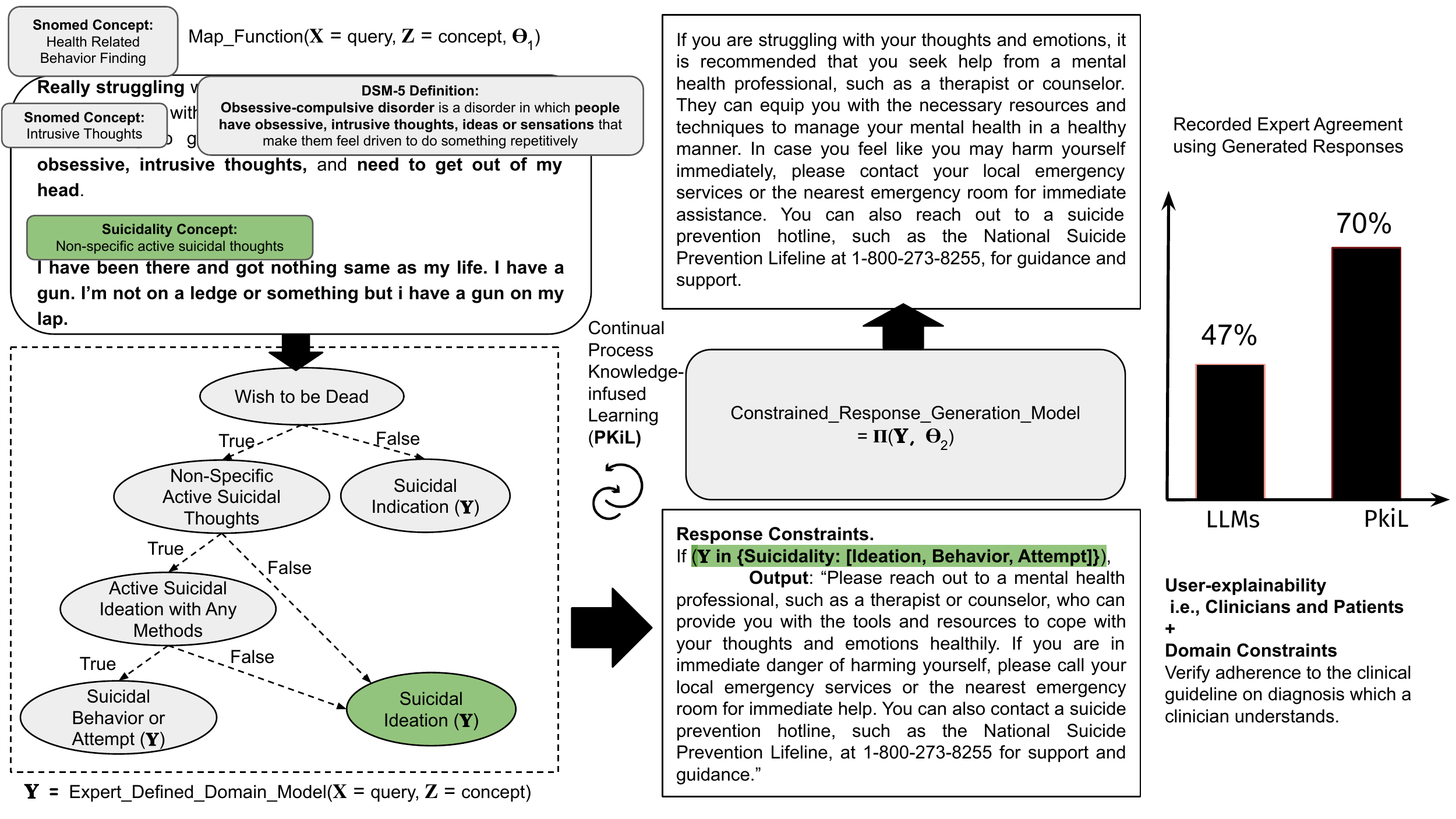}}
    \caption{\footnotesize depicts a pipeline that is fully differentiable from end to end. It consists of a composition of functions corresponding to various pipeline components. This pipeline enables the development of application-tailored AI systems that can be easily trained end-to-end. To accomplish this, trainable map functions are applied to raw data, converting it to concepts in the domain model. The example given in the figure relates to mental health diagnosis and conversational assistance. The map functions link fragments of raw data to decision variables in the diagnosis model, which are then used to apply constraints to the patient's response generated by the text generation model. Results from an existing implementation demonstrates that expert satisfaction levels reached 70\% using such a pipeline, compared to 47\% with LLMs in federated pipelines, such as OpenAI's text-Davinci-003 \cite{roy2022process}.}
    \label{fig:two_b}
\end{figure*}

\textbf{Application-Level Analysis of Methods in Category 2.}

For the systems belonging to category 2(a), tracing their chain-of-thought during processing immensely enhances the application-level aspects of user-explainability. However, the language model's ability to parse the input query and relate it to domain model concepts during response generation limits this ability ((\textbf{M}) in Figure \ref{fig:landscape}). Furthermore, the specification of domain constraints in natural language using prompt templates also limits the constraint modeling capability, which depends on the language model's ability to comprehend application or domain-specific concepts ((\textbf{M}) in Figure \ref{fig:landscape}). Federated pipelines excel in scalability since language models and application plugins that facilitate their use for domain-specific use cases are becoming more widely available and accessible ((\textbf{H}) in Figure \ref{fig:landscape}). Unfortunately, language models require an enormous amount of time and space resources to train, and hence continual domain adaptation using federated pipelines remains challenging ((\textbf{L}) in Figure \ref{fig:landscape}). Nonetheless, advancements in language modeling architectures that support continual learning goals are fast gaining traction. Category 2(b) methods show significant promise as  they score highly regarding all application-level aspects, including user-explainability, domain constraints, scalability across use cases, and support for continual adaptation to application-specific changes ((\textbf{H}) in Figure \ref{fig:landscape}). This is due to the high modeling flexibility and closely intertwined coupling of system components. Thus, a change in any particular component leads to positive changes in all components within the system's pipeline. Notably, in an implemented system for the mental health diagnostic assistance use case, shown in Figure \ref{fig:two_b}, we see drastic improvements in expert satisfaction with the system’s responses, further demonstrating the immense potential for 2(b) category methods.
\section{The Future of Neurosymbolic AI}
In this article, we compared different neurosymbolic architectures, considering their algorithm-level aspects, which encompass perception and cognition, and application-level aspects, such as user-explainability, domain constraint specification, scalability, and support for continual learning. The rapid improvement in language models suggests that they will achieve almost optimal performance levels for large-scale perception. Knowledge graphs are suitable for symbolic structures that bridge the cognition and perception aspects because they support real-world dynamism.  Unlike static and brittle symbolic logics, such as first-order logic, they are easy to update. In addition to their suitability for enterprise-use cases and established standards for portability, knowledge graphs are part of a mature ecosystem of algorithms that enable highly efficient graph management and querying. This scalability allows for modeling large and complex datasets with millions or billions of nodes.  

In summary, this article highlights the effectiveness of combining language models and knowledge graphs in current implementations. However, it also suggests that future knowledge graphs have the potential to model heterogeneous types of application and domain-level knowledge beyond schemas. This includes workflows, constraint specifications, and process structures, further enhancing the power and usefulness of neurosymbolic architectures. Combining such enhanced knowledge graphs with high-capacity neural networks would provide the end-user with an extremely high degree of algorithmic and application-level utility. The concern for safety is behind the recent push to hold further rollout of generative AI systems such as GPT*, since current systems could significantly harm individuals and society without additional guardrails.  We believe that guidelines, policy, and regulations can be encoded via extended forms of knowledge graphs such as shown in Figure \ref{fig:two_b} (and hence symbolic means), which in turn can provide explainability accountability, rigorous auditing capabilities, and safety. Encouragingly, progress is being made on all these fronts swiftly, and the future looks promising.
\section*{Acknowledgements}
This work was supported in part by the National Science Foundation under Grant 2133842, “EAGER: Advancing Neuro-symbolic AI with Deep Knowledge-infused Learning.”

\section*{Authors}

\noindent \textbf{Amit Sheth} is the founding director of the AI Institute of South Carolina (AIISC), NCR Chair, and a professor of Computer Science \& Engineering at USC. He received the 2023 IEEE-CS Wallace McDowell award and is a fellow of IEEE, AAAI, AAIA, AAAS, and ACM. Contact him at: \texttt{amit@sc.edu} 

\noindent \textbf{Kaushik Roy}
 is a Ph.D. student with an active publication record in the area of this article. Contact him at: \texttt{kaushikr@email.sc.edu} 

\noindent \textbf{Manas Gaur} is an assistant professor at UMBC. His dissertation was on Knowledge-infused Learning, with ongoing research focused on interpretability, explainability, and safety of the systems discussed in this article. He is a recipient of the EPSRC-UKRI Fellowship, Data Science for Social Good Fellowship, and AI for Social Good Fellowship, and was recently recognized as 2023 AAAI New Faculty. Contact him at: \texttt{manas@umbc.edu}
\bibliographystyle{IEEEtran}
\bibliography{references.bib}
\end{document}